\documentclass[10pt,twocolumn,letterpaper]{article}

\usepackage{cvpr}              %
\usepackage{graphicx}
\usepackage{amsmath}
\usepackage{amssymb}
\usepackage{booktabs}

\usepackage[pagebackref,breaklinks,colorlinks]{hyperref}
\usepackage[capitalize]{cleveref}

\usepackage{tikz}
\usepackage{comment}
\usepackage{amsmath,amssymb} %
\usepackage{wrapfig} %
\usepackage{color}

\usepackage{mathtools}
\usepackage[accsupp]{axessibility}  %

\usepackage[utf8]{inputenc} %
\usepackage[T1]{fontenc}    %
\usepackage{xr-hyper}
\usepackage{url}            %
\usepackage{booktabs}       %
\usepackage{xcolor,colortbl}         %
\usepackage{caption}

\usepackage{multirow}
\usepackage{enumitem}

\usepackage[normalem]{ulem}

\usepackage{subcaption}
\usepackage{float}
\usepackage{lscape}     

\usepackage{xspace}
\usepackage{setspace}
\usepackage{pifont}
\usepackage{comment}

\usepackage{enumitem}%

\crefname{section}{Sec.}{Secs.}
\Crefname{section}{Section}{Sections}
\Crefname{table}{Table}{Tables}
\crefname{table}{Tab.}{Tabs.}

\begin{document}

\title{Gaussian Shadow Casting for Neural Characters
}

\author{Luis Bolanos \qquad Shih-Yang Su \qquad Helge Rhodin\\
The University of British Columbia
}

\newcommand{\TODO}[1]{\textcolor{red}{TODO: #1}}
\newcommand{\todo}[1]{\textcolor{red}{#1}}
\newcommand{\sy}[1]{{\color{magenta}#1}}
\newcommand{\SY}[1]{{\color{magenta}(Shih-Yang: #1)}}
\newcommand{\LB}[1]{{\color{blue}(Luis: #1)}}
\newcommand{\lb}[1]{{\color{blue}#1}}
\newcommand{\hr}[1]{{\color{purple}#1}}
\newcommand{\HR}[1]{{\color{purple}(Helge: #1)}}

\newcommand{\tbf}[1]{\textbf{#1}}
\newcommand{\topic}[1]{\textbf{#1}}
\newcommand{\figref}[1]{Figure~\ref{#1}}
\newcommand{\secref}[1]{Section~\ref{#1}}
\newcommand{\feqref}[1]{Equation~\ref{#1}}
\newcommand{\tabref}[1]{Table~\ref{#1}}

\newcommand{\app}{appendix} %

\newcommand{\parag}[1]{\noindent\textbf{#1}}

\newcommand{\new}[1]{{\color{red}{#1}}}
\newcommand{\old}[1]{\textcolor{red}{\sout{#1}}}

\newcommand{\R}{\mathbb{R}}

\newcommand{\va}{\mathbf{a}}
\newcommand{\vb}{\mathbf{b}}
\newcommand{\vc}{\mathbf{c}}
\newcommand{\vd}{\mathbf{d}}
\newcommand{\ve}{\mathbf{e}}
\newcommand{\vf}{\mathbf{f}}
\newcommand{\vg}{\mathbf{g}}
\newcommand{\vh}{\mathbf{h}}
\newcommand{\vi}{\mathbf{i}}
\newcommand{\vj}{\mathbf{j}}
\newcommand{\vk}{\mathbf{k}}
\newcommand{\vl}{\mathbf{l}}
\newcommand{\vm}{\mathbf{m}}
\newcommand{\vn}{\mathbf{n}}
\newcommand{\vo}{\mathbf{o}}
\newcommand{\vp}{\mathbf{p}}
\newcommand{\vq}{\mathbf{q}}
\newcommand{\vr}{\mathbf{r}}
\newcommand{\vt}{\mathbf{t}}
\newcommand{\vu}{\mathbf{u}}
\newcommand{\vv}{\mathbf{v}}
\newcommand{\vw}{\mathbf{w}}
\newcommand{\vx}{\mathbf{x}}
\newcommand{\vy}{\mathbf{y}}
\newcommand{\vz}{\mathbf{z}}

\newcommand{\mA}{\mathbf{A}}
\newcommand{\mB}{\mathbf{B}}
\newcommand{\mC}{\mathbf{C}}
\newcommand{\mD}{\mathbf{D}}
\newcommand{\mE}{\mathbf{E}}
\newcommand{\mF}{\mathbf{F}}
\newcommand{\mG}{\mathbf{G}}
\newcommand{\mH}{\mathbf{H}}
\newcommand{\mI}{\mathbf{I}}
\newcommand{\mJ}{\mathbf{J}}
\newcommand{\mK}{\mathbf{K}}
\newcommand{\mL}{\mathbf{L}}
\newcommand{\mM}{\mathbf{M}}
\newcommand{\mN}{\mathbf{N}}
\newcommand{\mO}{\mathbf{O}}
\newcommand{\mP}{\mathbf{P}}
\newcommand{\mQ}{\mathbf{Q}}
\newcommand{\mR}{\mathbf{R}}
\newcommand{\mS}{\mathbf{S}}
\newcommand{\mT}{\mathbf{T}}
\newcommand{\mU}{\mathbf{U}}
\newcommand{\mV}{\mathbf{V}}
\newcommand{\mW}{\mathbf{W}}
\newcommand{\mX}{\mathbf{X}}
\newcommand{\mY}{\mathbf{Y}}
\newcommand{\mZ}{\mathbf{Z}}

\newcommand{\cA}{\mathcal A}
\newcommand{\cB}{\mathcal B}
\newcommand{\cC}{\mathcal C}
\newcommand{\cD}{\mathcal D}
\newcommand{\cE}{\mathcal E}
\newcommand{\cF}{\mathcal F}
\newcommand{\cG}{\mathcal G}
\newcommand{\cH}{\mathcal H}
\newcommand{\cI}{\mathcal I}
\newcommand{\cJ}{\mathcal J}
\newcommand{\cK}{\mathcal K}
\newcommand{\cL}{\mathcal L}
\newcommand{\cM}{\mathcal M}
\newcommand{\cN}{\mathcal N}
\newcommand{\cO}{\mathcal O}
\newcommand{\cP}{\mathcal P}
\newcommand{\cQ}{\mathcal Q}
\newcommand{\cR}{\mathcal R}
\newcommand{\cS}{\mathcal S}
\newcommand{\cT}{\mathcal T}
\newcommand{\cU}{\mathcal U}
\newcommand{\cV}{\mathcal V}
\newcommand{\cW}{\mathcal W}
\newcommand{\cX}{\mathcal X}
\newcommand{\cY}{\mathcal Y}
\newcommand{\cZ}{\mathcal Z}

\newcommand{\bR}{\mathbb{R}}
\newcommand{\mx}{\mathbf{x}}
\newcommand{\mj}{\mathbf{j}}
\newcommand{\mb}{\mathbf{b}}
\newcommand{\vmu}{\mathbf{\mu}}

\newcommand{\Irradiance}{\mI_i}
\newcommand{\mAlbedo}{\mA}
\newcommand{\mDepth}{\mD}
\newcommand{\mShadow}{\mS}
\newcommand{\mDiffuse}{\vc_d}
\newcommand{\mAmbient}{\vc_a}
\newcommand{\mLight}{\vn_l}
\newcommand{\mRGB}{\mathbf{RGB}}
\newcommand{\mRGBlit}{\mathbf{RGB_{lit}}}
\newcommand{\normal}{\vn_s} %

\newcommand{\irradiance}{\mI_i}
\newcommand{\albedo}{\mathbf{a}}
\newcommand{\depth}{\mathbf{d}}
\newcommand{\shadow}{s}
\newcommand{\surfacenormal}{\hat{\mathbf{n}}}
\newcommand{\lightcolor}{\mathbf{L_\text{col}}}
\newcommand{\ambient}{\mathbf{\hat{L}_\text{amb}}}
\newcommand{\lightdirection}{\mathbf{\hat{L}_\text{dir}}}
\newcommand{\pixelcoltrain}{\mathbf{c}}
\newcommand{\pixelcol}{\hat{\pixelcoltrain}}
\newcommand{\RGBloss}{\mathcal{L}_{\text{RGB}}}
\newcommand{\Maskloss}{\mathcal{L}_{\text{mask}}}
\newcommand{\Ambientloss}{\mathcal{L}_{\text{amb}}}
\newcommand{\Eikonalloss}{\mathcal{L}_{\text{Eikonal}}}
\newcommand{\Curvatureloss}{\mathcal{L}_{\text{Curvature}}}
\newcommand{\Greyloss}{\mathcal{L}_{\text{grey}}}
\newcommand{\GaussianDloss}{\mathcal{L}_{\text{gDensity}}}
\newcommand{\GaussianUloss}{\mathcal{L}_{\text{gMean}}}
\newcommand{\GaussianSloss}{\mathcal{L}_{\text{gSigma}}}
\newcommand{\mask}{\rho}
\newcommand{\accumulation}{\hat{\rho}}

\newcommand{\dotprod}{\boldsymbol{\cdot}}

\definecolor{Gray}{gray}{0.85}

\definecolor{DeepGreen}{rgb}{0.15,0.60,0.15}
\newcommand{\canon}[1]{#1^c}
\newcommand{\world}[1]{#1^w}
\newcommand{\Ltwo}[1]{\vert\vert #1\vert\vert^2_2}

\newcommand{\obs}[1]{#1^\vo}

\twocolumn[{
\renewcommand\twocolumn[1][]{#1}
\maketitle
\begin{center}%
    \centering%
    \captionsetup{type=figure}%
    \includegraphics[width=0.99\linewidth]{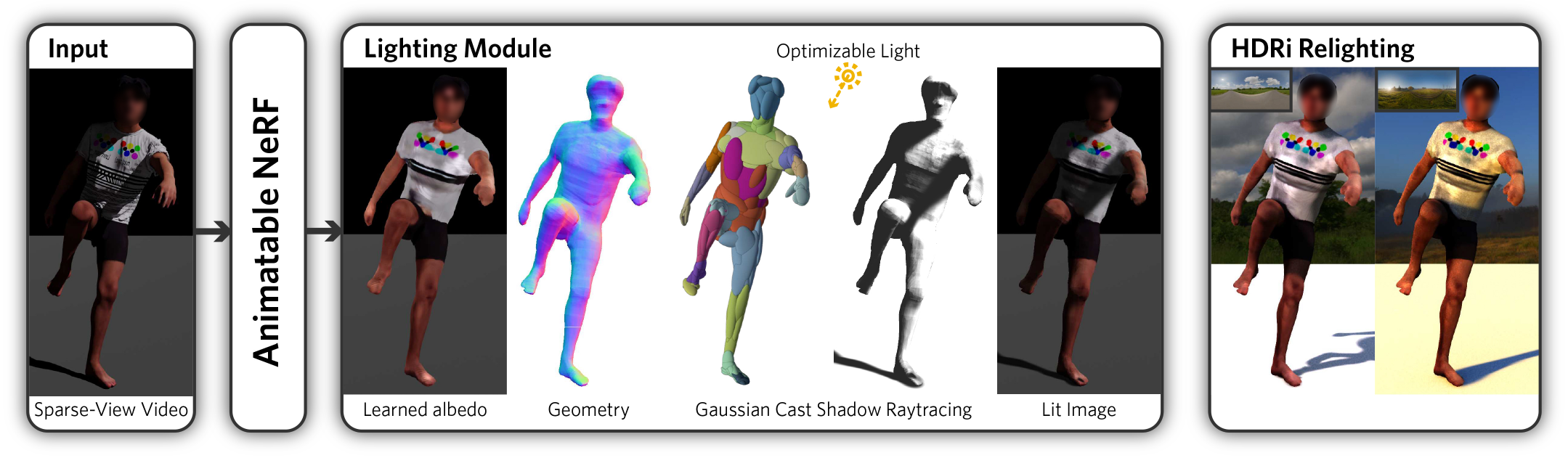}%
    \captionof{figure}{
    \textbf{Gaussian Shadow Casting (GSC):} Our method is able to reconstruct 3D neural characters from a sparse set of videos in settings with strong directional illumination. GSC uses a sum of Gaussians density model to cast secondary shadow rays efficiently with an analytic formula. Our method learns to remove shadows from the neural color field, allowing us to relight in novel illuminations.
    \color{red}~All faces are blurred for anonymity.
    }
\label{fig:teaser}%
\end{center}%

}]

\newlength\tindent
\setlength{\tindent}{\parindent}
\noindent\begin{minipage}{1\linewidth}%
\begin{abstract}
\setlength{\parindent}{\tindent}
\indent Neural character models can now reconstruct detailed geometry and texture from video, but they lack explicit shadows and shading, leading to artifacts when generating novel views and poses or during relighting. 
It is particularly difficult to include shadows as they are a global effect and the required casting of secondary rays is costly.
We propose a new shadow model using a Gaussian density proxy that replaces sampling with a simple analytic formula. It supports dynamic motion and is tailored for shadow computation, thereby avoiding the affine projection approximation and sorting required by the closely related Gaussian splatting.  
Combined with a deferred neural rendering model, our Gaussian shadows enable Lambertian shading and shadow casting with minimal overhead.
We demonstrate improved reconstructions, with better separation of albedo, shading, and shadows in challenging outdoor scenes with direct sun light and hard shadows. Our method is able to optimize the light direction without any input from the user.
As a result, novel poses have fewer shadow artifacts and relighting in novel scenes is more realistic compared to the state-of-the-art methods,
providing new ways to pose neural characters in novel environments, increasing their applicability.
\end{abstract}
\end{minipage}

\section{Introduction}

It is now possible to reconstruct animatable 3D neural avatars from video but methods
do not account for accurate lighting and shadows. 
They have to rely on recordings that have soft uniform lighting, which precludes recording outdoors in direct sun light and on film sets with spotlights, %
and most are unable to relight characters in novel environments, limiting their applicability in content creation. 

The most recent body models \cite{jiang2022neuman, kwon2021neural, peng2021neural, su2022danbo, su2023npc, wang2022arah} which are based on neural radiance fields (NeRFs) \cite{mildenhall2020nerf}, approximate the light transport by casting primary rays between the camera and the scene, sampling the underlying neural network dozens of times along each ray to obtain the density and color. As they do not include an illumination model, the color that the NeRF learns includes lighting, shadow, and view-dependent effects. 
Learning a body model in a challenging scene with a strong directional light source, such as the sun, leads to the neural field overfitting to the observed shadows. It does not generalize to novel poses, as the cast shadows are global effects where movement of a joint could affect the appearance of other distant areas of the body. \figref{fig:teaser} shows such setting. This is in contrast to local shading effects such as wrinkles in clothing which current body models can successfully reconstruct.

To cast shadows within NeRF, secondary ray tracing from the reconstructed body model to the light source is an option. Although the predominant NeRF formulation enables casting secondary rays without change, it comes with a massive computational cost. For each sample along the primary ray, an equal number of secondary rays would have to be computed, each with multiple samples, leading to a quadratic, instead of linear, complexity in the number of samples per pixel. As a result, current re-lighting models only support diffuse reflection \cite{chen2022relighting}, hard shadows that do not generalize to novel poses \cite{iqbal2022rana}, and soft dynamic shadow maps by approximate sphere tracing~\cite{xu2023relightable}.

Our core contribution is introducing an additional volumetric density field that is 
approximate but significantly speeds up dynamic shadow casting while still maintaining differentiability and smoothness for gradient-based optimization. We introduce an anisotropic Gaussian density model and associated rendering functions that approximate the fine-grained density of the NeRF. The Gaussians have the beneficial property that we can integrate their density along a ray in closed form, thereby avoiding any sampling steps. 
Our derivation and implementation differs significantly from existing work using Gaussians for rendering. 
Compared to Gaussian Splatting \cite{sridhar2014real,rhodin2015visibility,kerbl3Dgaussians}, we neither require an affine approximation nor back-to-front ordering.
Compared to Gaussian density models we alleviate their sampling \cite{rhodin2015visibility} with an analytic integration and extend the existing analytic integration~\cite{rhodin2016contour} to apply to anisotropic Gaussians.
Notably, the Gaussian density is optimized alongside the NeRF without requiring a reference mesh such as SMPL \cite{loper2015smpl}; it is template-free.

To further reduce runtime, we use a deferred shading approach \cite{chen2022relighting} in which the first rendering pass computes the albedo, depth, and normal for each pixel. The second pass casts only one secondary ray per pixel from the estimated surface point to the light source. This makes shadow computations independent of the number of samples in the NeRF, avoiding the mentioned quadratic complexity.

Our experiments with strong directional light and cast shadows demonstrate that our explicit lighting reduces the occurrence of artifacts in novel-view and novel-pose synthesis tasks. 
\figref{fig:teaser} shows how our method is able to disentangle lighting and shadows from the avatar's albedo given sparse-view data from only a single illumination. We take advantage of the dynamic aspect of the data where we can observe the same body part in multiple illuminations as the subject moves. We further demonstrate the ability to optimize the unknown light directions without any user input or careful initialization. Moreover, relighting of the neural character enables us to composite recorded motions into novel scenes realistically, making them directly applicable in computer graphics and entertainment industries, as demonstrated by the HDRi re-lighting in \figref{fig:teaser}-right.

\section{Related Work}

We build on neural body models using NeRF~\cite{mildenhall2020nerf}, which we introduce briefly. 
The subsequent discussion focuses on relighting methods for 3D scenes and body models as well as how Gaussians are used in rendering and reconstruction.

\parag{Neural avatars} %
model dynamic performances by conditioning the neural rendering model on a template mesh driven by skeleton motion \cite{bagautdinov2021driving, iqbal2022rana, li2023posevocab, liu2021neuralactor, kwon2021neural, wang2022arah, zheng2022structured, zheng2023avatarrex} or template-free by linking neural fields directly to a skeleton \cite{li2022tava, noguchi2021narf, su2021anerf, su2022danbo, su2023npc}.
Our implementation uses the more flexible template-free approach but it is general enough to extend to any NeRF-based model.

\parag{Static NeRF scene relighting} approaches
can be categorized by either implicit \cite{rudnev2022nerfosr, srinivasan2021nerv, derksen2021snerf, boss2021nerd, zhang2021nerfactor} or explicit \cite{wang2023fegr, guo2020objectcentric} implementations. In implicit methods, the NeRF's MLP is extended to further output illumination data such as shadow, direct and indirect illumination or occlusion maps \cite{rudnev2022nerfosr, srinivasan2021nerv, derksen2021snerf}, or decompose the scene into material properties such as metallicity and roughness which can be used in a Bidirectional Reflectance Distribution Function (BRDF) lighting model \cite{boss2021nerd, zhang2021nerfactor}. These extended MLPs are conditioned at training and test time on lighting information such as spherical harmonics coefficients \cite{rudnev2022nerfosr}, or light direction \cite{derksen2021snerf}. Implicit methods require large amounts of data in both multi-view and multiple illuminations with lighting information known or estimated \cite{rudnev2022nerfosr, derksen2021snerf}. 
Explicit methods simulate how real light interacts with the environment which improves the generalizability to novel illuminations but are difficult to extend to dynamic scenes or objects. These methods either utilize a secondary data structure such as proxy geometries where lighting computations can be done using established methods \cite{wang2023fegr}, or attempt to cast the necessary secondary rays within the neural field's volume which comes with a significant computational burden \cite{guo2020objectcentric}.

\parag{Dynamic Neural character relighting}
has been built on top of volume rendering methods \cite{ranjan2023facelit, zhou2023relightable, xu2023relightable, yang2023travatar, li2023megane, bi2021faces, chen2022relighting} as well as 2D CNN based models \cite{iqbal2022rana}.
Implicit methods 
again require large amounts of data which can only be captured using light stages with known illumination \cite{zhou2023relightable, yang2023travatar, li2023megane, bi2021faces} or, across multiple subjects for faces that are self-similar, each captured in a different setting with in-the-wild illumination~\cite{ranjan2023facelit}.
Our model provides dynamic and explicit shading and is most closely related to the following three methods.

RANA \cite{iqbal2022rana} uses SMPL+D \cite{alldieck2019smpld} to estimate the coarse geometry of a person and extract an albedo texture map using TextureNet \cite{huang2019texturenet}. Given a target pose, they render person-specific neural features alongside coarse albedo and normals from the SMPL-D model. These are passed through two CNNs to refine the albedo and normal maps. Finally, they generate a light map using spherical harmonics and the normal map which is multiplied by the albedo map to obtain the final lit image. While spherical harmonics allow a wide array of lighting conditions to be simulated, cast shadows are not present, e.g., an arm casting a shadow on the body.
Our work implements a Gaussian density model \cite{rhodin2015visibility, rhodin2016contour} to facilitate fast and efficient \emph{secondary} ray tracing to compute these cast shadows.

Likewise, Relighting4D \cite{chen2022relighting} uses SMPL \cite{loper2015smpl} to condition a 4D neural field of latent features which are trilinearly interpolated based on the nearby vertices to the query location. The latent features are passed through an MLP to obtain geometry, occlusion, and reflectance properties which are fed through a BRDF to get the final lit image. It is able to estimate the light probe, and at inference time, be able to switch the light probe to a new illumination. However, Relighting4D was not designed to work with hard shadows in novel poses, which is the focus point of our work.

Finally, Xu et al. \cite{xu2023relightable} utilize a signed distance field (SDF) based approach to learn a neural human avatar which utilizes SMPL-based inverse Linear Blender Skinning (LBS) and a displacement field to obtain canonical features. They utilize Hierarchical Distance Queries (HDQ) to compute minimum distances from world space to surface locations and perform sphere tracing to obtain material and surface properties for each camera ray. They further take advantage of HDQ through the SDF by computing soft visibility maps towards a learned light probe. While HDQ allows for fast occlusion checks, their solution focuses on soft approximate shadows whereas our work enables hard shadow casting.

\parag{Gaussians}
have been used in rendering applications as differentiable methods for computing visibility and occlusions \cite{rhodin2015visibility, rhodin2016contour, stoll2011fast}, as components of environment maps \cite{zhang2021physg}, or as a means to improve rendering efficiency for neural scenes \cite{kerbl3Dgaussians}. Most methods are limited to spherical Gaussians~\cite{rhodin2015visibility, rhodin2016contour}, while
Gaussian Splatting uses an affine approximation that is only accurate when many small Gaussians are used~\cite{kerbl3Dgaussians}, and Sridhar et al.~use an approximation by perspective projection of ellipsoids~\cite{sridhar2014real}.
Our work extends Rhodin et al.~\cite{rhodin2015visibility, rhodin2016contour} to use anisotropic Gaussians, without introducing any approximation, and tailors the analytic formulas and implementation towards shadow casting. 

\begin{figure*}[t!]
\centering
\includegraphics[width=0.8\linewidth]{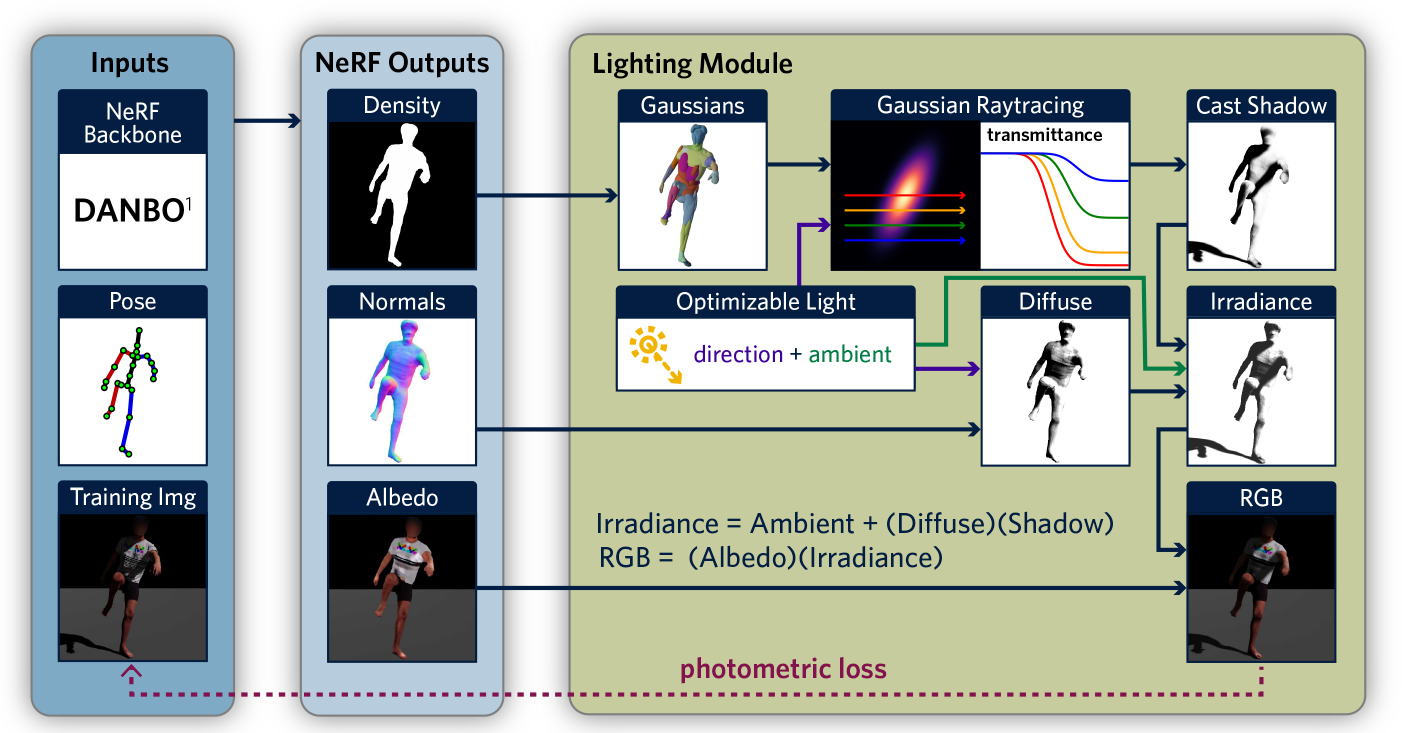}
\caption{\textbf{Method Overview.} 
Our method takes as input images and poses of a person. Using a neural radiance field as a backbone$^{1}$\cite{su2022danbo}, density, normals, and albedo values are volumetrically reconstructed and rendered. We fit a sum of 3D anisotropic Gaussian density model to approximate the neural density field and compute shadow maps using our novel anisotropic Gaussian ray occlusion equations. The shadow map is combined with a diffuse shading pass to produce the lit image. The whole model is optimized with a photometric loss against the training images. Our method is able to optimize the light direction and ambient intensity without any initialization. It also separates albedo from shading and shadow, allowing us to relight the model.
}
\label{fig:method}
\end{figure*}

\section{Method}
Our method reconstructs a neural character from a set of $N$ images of width $W$ and height $H$, $\{\mI_t \in \R^{H \times W \times 3} \}_{t=1}^N$, and corresponding character poses $\theta_t \in \R^{J \times 4 \times 4}$. The pose is represented as a skeleton with one $4 \times 4$ local-to-world transformation matrix for each of the $J$ joints.
\figref{fig:method} gives an overview of our method.
A key element of our design is a deferred illumination model~\cite{thies2019deferred} that separates the rendering into computing albedo, $\albedo \in \R^3$, surface normal, $\surfacenormal \in \R^3$, and depth, $\depth \in \R$, in a first pass and subsequently adding shading and shadow, $\shadow \in [0,1]$, in a second pass. 
Our key contribution is the closed form formula for the shadow~$\shadow$.

\subsection{Deferred Neural Illumination}
\label{sec:deferred}
Our volumetric body model is optimized on a reconstruction objective, $\RGBloss$ that minimizes the squared difference between the input images $\mI_t$ and the rendering of the model.
We test our method using DANBO \cite{su2022danbo}. It outputs a color and density for samples $\vx$ along the primary view rays. These are subsequently integrated to compute a color, which we interpret as the albedo, $\albedo$. 
The illuminated color for a given pixel of the reconstructed image, $\pixelcol$, is computed by a Lambertian reflectance model,
\begin{equation}
\pixelcol = \albedo(\theta_t) \left(\ambient + \shadow(\theta_t) \lightcolor (\lightdirection \dotprod \surfacenormal(\theta_t)) \right).
\label{eq:lambertianshading}
\end{equation}
This diffuse shading model illuminates the entire body 
with an ambient light $\ambient$ and a directional light with color $\lightcolor$. 
The directional light intensity is attenuated by the cast shadows $\shadow$ and the cosine angle between the surface normal $\surfacenormal$ and light direction $\lightdirection$.

\paragraph{Shading extensions.} The benefit of the deferred rendering approach is that it lets us compute lighting information only once for each pixel, as opposed to at every sample location of the volumetric ray tracing leading to significantly faster computation.
To be applicable, we extend DANBO to yield surface normals $\surfacenormal$ and depth $\depth$ for a given view ray. The former we attain by switching the density formulation to a signed distance function with an Eikonal loss. The normal is then readily estimated
by differentiating the distance with respect to the original query location $\vx$ as in~\cite{yariv2021volume}.
We compute $\depth$ likewise to albedo $\albedo$ by integrating the sample's $\vx$ positions along the ray, weighted by their density and transmittance.
Furthermore, we fix the intensity of the directional light to white with a magnitude of $1.5$. Without fixing the directional light intensity, the equation would be over parametrized and lead to ambiguities.

\subsection{Gaussian Shadow Casting}
\label{sec:shadow}
For the sake of modeling shadows more efficiently, we represent the body shape with a set of Gaussians rigidly attached to the skeleton model. The relative positions, orientation, and size of the Gaussians are optimized to approximate the density of the neural field and to allow for a fast, efficient, and closed-form solution for integration along a ray (occlusion checking).
Our model extends previous work \cite{rhodin2015visibility,rhodin2016contour} by using anisotropic Gaussians (variable scale and rotation along each axis) and avoids the need for sampling during integration as in NeRF.

\begin{figure}[h]
\centering
\includegraphics[width=1.0\linewidth]{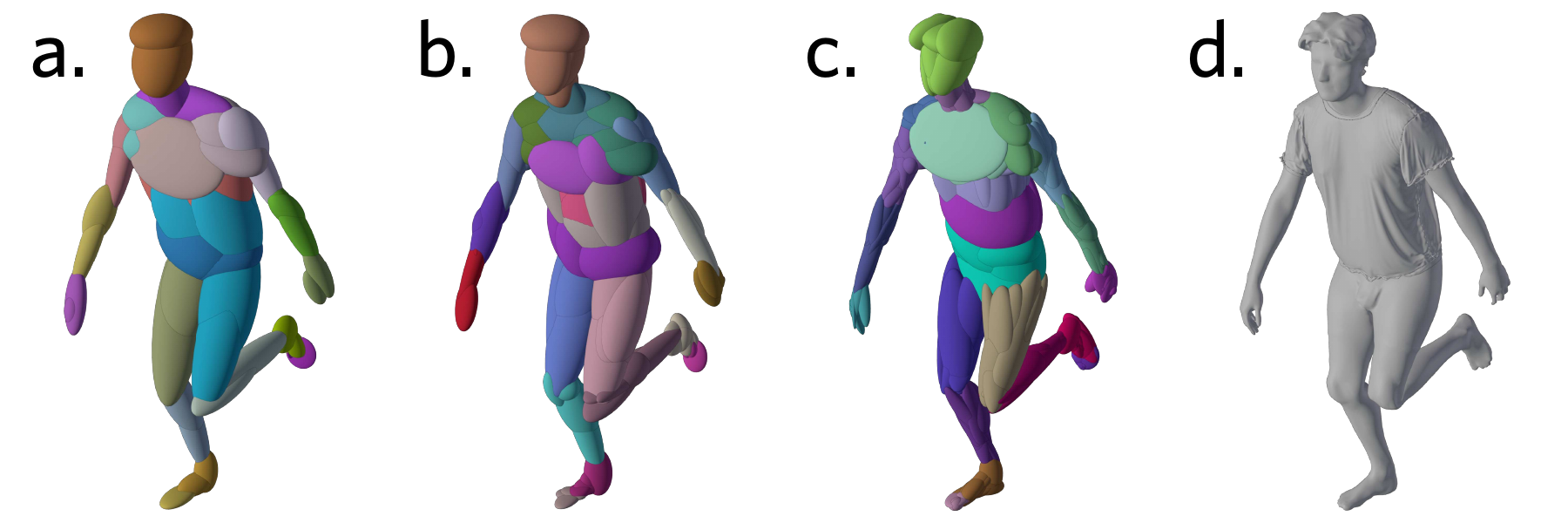}
\caption{\textbf{Gaussian Density Model.} 
The approximation to the NeRF's density field using a sum of 3D anisotropic Gaussians using: a) 2 Gaussians per bone, b) 4 Gaussian per bone, and c) 8 Gaussian per bone; d) is the groundtruth mesh. \emph{Note: ellipses are scaled to 2.5 STD of the Gaussians (99$^{\text{th}}$ percentile})
}
\label{fig:gvm}
\end{figure}

\paragraph{Anisotropic Gaussian body model.}
We define the anisotropic Gaussian density model as the matrix $\mG \in \mathbb{R}^{J \times K \times 13}$, with 
$K$ being the number of Gaussians per joint, typically $\sim{8}$, and the columns representing the 3D mean ($\mu_x, \mu_y, \mu_z$), the axis aligned standard deviations ($\sigma_x, \sigma_y, \sigma_z$), the rotation defined using the 6 DOF representation ($R_{0,0}, R_{0,1}, R_{0,2}, R_{1,0}, R_{1,1}, R_{1,2}$) \cite{zhou2019continuity}, and density ($\cC$). \figref{fig:gvm} gives examples with varying numbers of Gaussians.

The 3D density function, $\mG(\vx)$, defines the density of the Gaussian model at the query location $\vx$ in world space. We define the density function of a single 3D anisotropic Gaussian as
\begin{equation}
\mG_{i}(\vx) = \cC\exp{\left[-0.5\left((\vmu-\vx)^T)\Sigma^{-1}(\vmu-\vx)\right)\right]},
\end{equation}
where the precision matrix $\Sigma^{-1} = \mR^T\mD\mR$ and $\mR$ is the rotation matrix computed from the 6 DOF representation and $\mD = \text{diagonal}(1/\sigma_x^2, 1/\sigma_y^2, 1/\sigma_z^2)$.

The density of the entire Gaussian model is the sum of the density of each. The query location is transformed to the local space of the given Gaussian's joint $j$ at time-step $t$ using the world-to-local transformation matrix $\theta_{t,j}^{-1}$, rigidly attaching the Gaussians to the underlying skeleton and facilitating animation,
\begin{equation}
\mG(\vx) = \sum_{i=0}^{J \times K}\mG_i(\theta_{t,j}^{-1}\vx).
\end{equation} %
We jointly fit the parameters of the Gaussian density model to the neural field by minimizing  $\GaussianDloss$, the L2 error between the density function $\mG(\vx)$ and the target neural density field at query location $\vx$. We detach the gradients of the neural density field to optimize the Gaussians, fitting the Gaussians to the neural field and not the other way around.

\paragraph{Gaussian Ray Tracing.}
\label{sec:gaussianray}

\begin{figure}[t]
\centering
\includegraphics[width=1.0\linewidth]{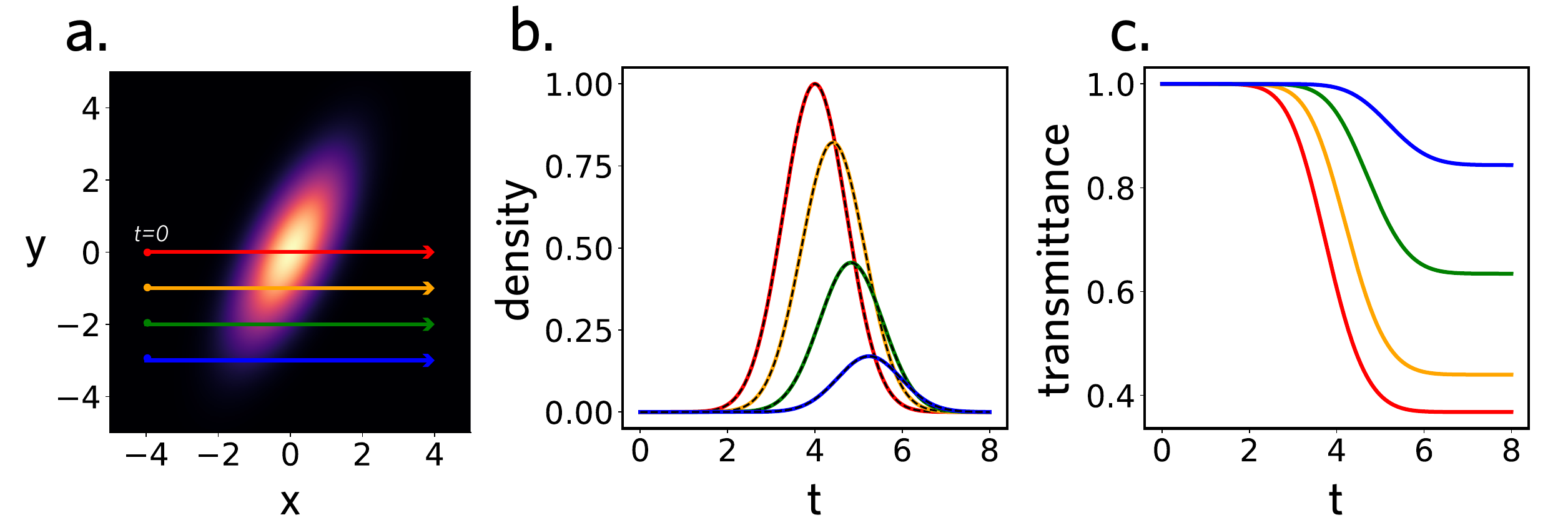}
\caption{\textbf{3D Anisotropic Gaussian Raytracing.} 
a) A cross-section of a 3D anisotropic Gaussian with rays passing through the Gaussian. b) The computed 1D Gaussians resulting from our derivation in~\secref{sec:gaussianray} (colored solid), compared to sampling the 3D Gaussian directly (dashed), with their exact match validating the correctness. c) The transmittance along each ray which is used as the shadow map value.
}
\label{fig:gaussianraytracing}
\end{figure}

\figref{fig:gaussianraytracing} shows how casting a ray, $r$ with ray origin $\vr_o \in \R^{3\times1}$ and direction $\vr_d \in \R^{3\times1}$, through a 3D anisotropic Gaussian results in a 1D Gaussian density along the ray. Through the Gaussian body model, this equates to a sum of 1D Gaussians for which analytic integrals can be computed. The amount of occlusion these rays experience is equal to the sum of the integrals of each of the 1D Gaussians across the rays. The transmittance value, $\cT$, used as the shadow map value, $s$, is the exponential of the negative integral from the start of the ray, $t=0$, to the length of the ray, $t=l$,
\begin{equation}
\label{eq:1dshadow}
\shadow = \cT_{r} = \exp\left[{-\sum_{i=0}^{J \times K}\int_{0}^{l} \mG_{i}^{r}}\right].
\end{equation}
$G_i^{r}$ is the 1D Gaussian created by the ray, $r$, going through the 3D anisotropic Gaussian, $G_i$ with mean $\mu\in\R^{3\times1}$ and precision matrix $\Sigma^{-1} \in \R^{3\times3}$. We derive in the supplemental how the 1D Gaussian's density function takes the form
\begin{equation}
G_i^{r^s} = \Bar{\cC} \cdot \exp{\left(-\frac{(\Bar{\mu}-x)^2}{2\Bar{\sigma}^2}\right)},
\end{equation}
where
\begin{align}
\Bar{\cC} &= \cC \exp{\left(-0.5\left((\mu - \vr_o)^T\Sigma^{-1}(\mu - \vr_o) - \frac{\Bar{\mu}^2}{\Bar{\sigma}^2}\right)\right)},
\nonumber\\%
\Bar{\mu} &= \frac{\vr_d^T\Sigma^{-1}(\mu-\vr_o)}{\vr_d^T\Sigma^{-1}\vr_d},\text{ and }
\nonumber\\%
\Bar{\sigma} &= \sqrt{\frac{1}{\vr_d^T\Sigma^{-1}\vr_d}}.
\label{eq:1ddensity}
\end{align}
This formula is more complex than in \cite{rhodin2016contour}, as it now accounts for anisotropic Gaussians with an arbitrary covariance instead of isotropic Gaussians. The comparison to sampling the 3D Gaussian in~\figref{fig:gaussianraytracing} validates their correctness. It lets us compute the cumulative density function (CDF) analytically, thereby avoiding the sampling in classical NeRFs,
\begin{equation}
\label{eq:1dintegral}
\int_{0}^{x} \mG_{i}^{r^s} = \Bar{\cC} \cdot 0.5 \cdot\left(1 + \text{erf}\left(\frac{x-\Bar{\mu}}{\Bar{\sigma}\sqrt{2}}\right)   \right).
\end{equation}
Together with~\feqref{eq:1dshadow}, this integral computes the shadow $s$ when applied to the secondary ray with origin $r_o^s$, as the point on the subject's surface computed from the depth map $\depth$.

\subsection{Optimization}

In addition to the introduced reconstruction loss $\RGBloss$, $\Eikonalloss$ for SDF regularization as in~\cite{gropp2020igr}, and Gaussian fitting $\GaussianDloss$, we regularize the training with i) a $\Maskloss = |\accumulation - \mask|$ that regularizes density by minimizing the difference between integrated accumulation, $\accumulation$, and the foreground mask, $\mask$, ii) $\Ambientloss = ||\ambient - 0.1||^2$ preferring small ambient light values, iii) $\GaussianSloss$ that prevent too large or small Gaussians, and iv) $\GaussianUloss$ that pulls Gaussians closer to the center of the bones.

Training proceeds in three stages. In stage I, the reconstruction loss is replaced with one that encourages predicting gray inside the silhouette, to learn a rough body shape without illumination effects. In stage II, $\GaussianDloss$ and its regularizers are introduced, allowing the Gaussian density model to fit. Finally, in stage III, the $\RGBloss$ takes over to optimize the light direction and learn the albedo. Additional training details are provided in the supplemental.

\begin{table*}[t]
\caption{\textbf{Novel-pose synthesis (all test frames).} Our Gaussian Shadow Casting model achieves consistently better PSNR scores for novel pose renderings as it properly models the hard shadows cast by the limbs in novel positions. Existing methods only shine in perceptual metrics (SSIM and LPIPS) as these normalize contrast and hence lessen the impact of proper shadows and shading.}
\centering
\resizebox{0.66\linewidth}{!}{
\setlength{\tabcolsep}{0pt}
\begin{tabular}{lcccccccccccccccccccccccc}
\toprule
& \multicolumn{3}{c}{Synthetic ($N=57$)} & \multicolumn{3}{c}{Real S1 ($N=200$)} & \multicolumn{3}{c}{Real S2 ($N=200$)} & \multicolumn{3}{c}{Average} \\
\cmidrule(lr){2-4}\cmidrule(lr){5-7}\cmidrule(lr){8-10}\cmidrule(lr){11-13}%
  & PSNR$\uparrow$  & SSIM$\uparrow$  & LPIPS$\downarrow$ 
  & ~PSNR\phantom{$\uparrow$}& SSIM\phantom{$\uparrow$}  & LPIPS\phantom{$\uparrow$} 
  & ~PSNR\phantom{$\uparrow$}& SSIM\phantom{$\uparrow$}  & LPIPS\phantom{$\uparrow$} 
  & ~PSNR\phantom{$\uparrow$}& SSIM\phantom{$\uparrow$}  & LPIPS\phantom{$\uparrow$} 
\\
\rowcolor{Gray}
DANBO \cite{su2022danbo}
& 17.52& 0.756& 0.195
& \underline{16.57}& \textbf{0.599}& \textbf{0.328}
& \underline{17.69}& \textbf{0.588}& \textbf{0.325}
& \underline{17.26}& \underline{0.648}& \textbf{0.283}\\
NPC \cite{su2023npc}
& 17.57& 0.758& 0.188
& 16.33& 0.590& 0.334
& 17.47& 0.575& 0.328
& 17.12& 0.641& \textbf{0.283}\\
\rowcolor{Gray}
Ours
& \textbf{22.04}& \textbf{0.829}& \textbf{0.166}
& \textbf{17.57}& \underline{0.592}& 0.356
& \textbf{18.29}& \underline{0.577}& 0.351
& \textbf{19.30}& \textbf{0.666}& \underline{0.291}\\
\end{tabular}
\label{tab:baselines-novel-pose}
}
\end{table*}
\begin{table*}[t]
\caption{\textbf{Novel-pose synthesis (subset of test set with observed self-casting shadows).} Our Gaussian Shadow Casting renders novel poses with strong hard shadows well. Our scores drop marginally on these hard frames compared to all frames in \cref{tab:baselines-novel-pose}, while the baselines drop significantly.}
\centering
\resizebox{0.666\linewidth}{!}{
\setlength{\tabcolsep}{0pt}
\begin{tabular}{lcccccccccccccccccccccccc}
\toprule
& \multicolumn{3}{c}{Synthetic ($N=15$)} & \multicolumn{3}{c}{Real S1 ($N=41$)} & \multicolumn{3}{c}{Real S2 ($N=36$)} & \multicolumn{3}{c}{Average} \\
\cmidrule(lr){2-4}\cmidrule(lr){5-7}\cmidrule(lr){8-10}\cmidrule(lr){11-13}%
  & PSNR$\uparrow$  & SSIM$\uparrow$  & LPIPS$\downarrow$ 
  & ~PSNR\phantom{$\uparrow$}& SSIM\phantom{$\uparrow$}  & LPIPS\phantom{$\uparrow$} 
  & ~PSNR\phantom{$\uparrow$}& SSIM\phantom{$\uparrow$}  & LPIPS\phantom{$\uparrow$} 
  & ~PSNR\phantom{$\uparrow$}& SSIM\phantom{$\uparrow$}  & LPIPS\phantom{$\uparrow$} 
\\
\rowcolor{Gray}
DANBO \cite{su2022danbo}
& 17.78& 0.740& 0.209
& 15.11& 0.559& \textbf{0.354}
& 16.55& \textbf{0.547}& \textbf{0.353}
& \underline{16.48}& \underline{0.615}& 0.305\\
NPC \cite{su2023npc}
& \underline{17.81}& \underline{0.741}& \underline{0.201}
& 14.88& 0.553& \underline{0.357}
& 16.51& 0.538& \underline{0.355}
& 16.40& 0.611& \underline{0.304}\\
\rowcolor{Gray}
Ours
& \textbf{22.13}& \textbf{0.821}& \textbf{0.175}
& \textbf{16.88}& \textbf{0.572}& 0.365
& \textbf{17.40}& \underline{0.544}& 0.371
& \textbf{18.81}& \textbf{0.646}& \textbf{0.303}\\
\end{tabular}
\label{tab:baselines-novel-pose-shadows}
}
\end{table*}
\begin{figure*}[t!]
\centering
\includegraphics[width=0.8\linewidth]{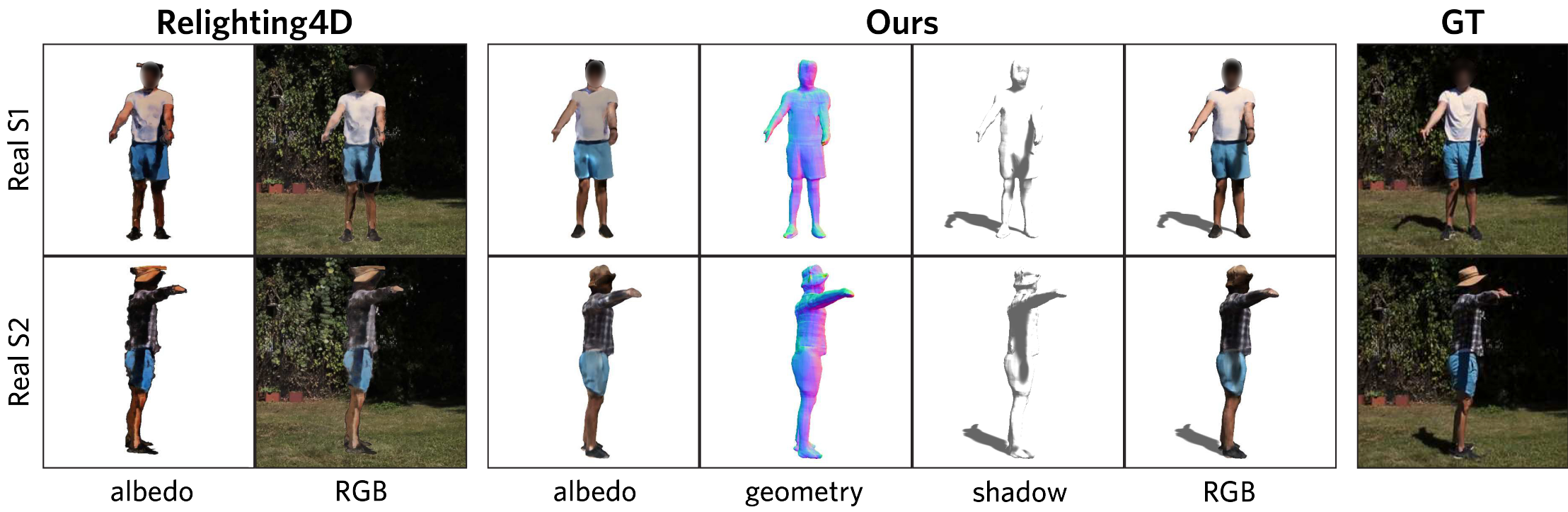}
\caption{\textbf{Qualitative Comparison (train).} 
Our method is able to better estimate albedo where shadows are not baked in as part of the neural field. We compare against Relighting4D on training images as their underlying neural body model is unable to handle novel poses.
}
\label{fig:qual_train}
\end{figure*}

\begin{figure*}[t!]
\centering
\includegraphics[width=1.0\linewidth]{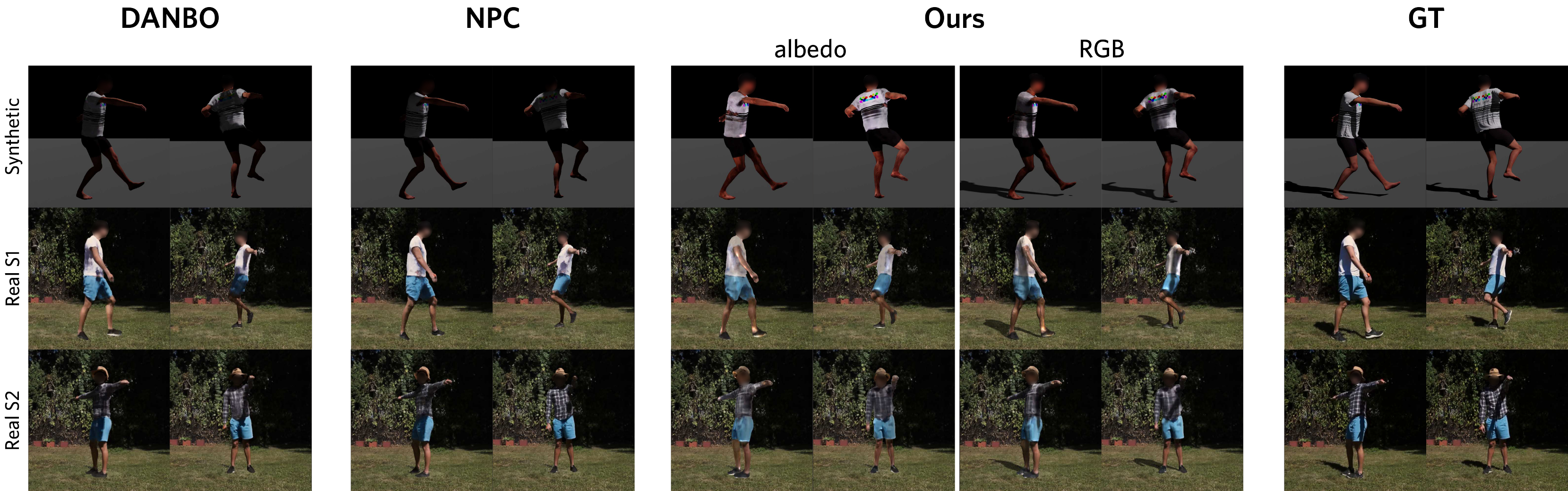}
\caption{\textbf{Outdoors in sunlight (test).} 
Our method can more accurately reproduce the shadow in novel poses compared to the baselines.
}
\label{fig:qual_baselines}
\end{figure*}

\section{Results}

We evaluate our method on synthetic sequences, as done in prior work~\cite{iqbal2022rana}. However, this does not test performance in real world conditions. Hence, we captured a new dataset in direct sunlight and compare to the most closely related baselines, showing significantly improved relightable body models.

The supplemental document provide additional qualitative comparisons, including relighting with HDRi environment maps.

\paragraph{Synthetic dataset.}
We obtained a textured mesh of a subject with uniform white illumination with a 3D full body scanner (VITUS 3D Body Scanner). A Blender \cite{blender} cloth simulation was applied to a shirt over the scan and the character was automatically rigged and animated using Mixamo \cite{mixamo}. We use the `swing-dance' animation as the driving motion as it contains a variety of poses from all body angles. 
Three cameras are placed  around the subject at 90 degrees from each other. 
A directional light source illuminates the scene with a slight ambient contribution such that the shadowed areas were not fully black. The animations are rendered using the Cycles render engine. In addition, ground-truth pose and segmentation masks are exported. We split the dataset into 600/57 images for the train/test sets, with the test set including 57 images with novel poses, out of which 15 have a strong hard shadow that we test separately.%

\paragraph{Outdoor sunlight dataset.}
We recorded two sequences of real human motion in an outdoor scene during a sunny day. We capture the data using 3 cameras (Canon EOS R8, Canon EOS 70D, iPhone12) and obtain SMPL estimates using EasyMocap \cite{easymocap,dong2020motion, dong2021fast}. Segmentation masks were obtained using the Segment Anything Model (SAM) \cite{kirillov2023segany}. We divide the frames into 600/200 images for the train/test splits, using all three cameras for training. %

\paragraph{Baselines}
We evaluate our method using the hard illumination dataset against Relighting4D \cite{chen2022relighting}. Due to code being unavailable and their original evaluation not testing direct shadow casting, we were not able to compare against RANA \cite{iqbal2022rana} and the shading approach by Xu et al. \cite{xu2023relightable}, which model illumination well but do not account for hard shadows. We do compare our work with other template-less neural body models~\cite{su2022danbo,su2023npc}, highlighting the drawback when not explicitly modeling lighting. DANBO \cite{su2022danbo} is our neural field backbone. NPC \cite{su2023npc} forms the current state-of-the-art template-less neural character model.

\subsection{Novel-pose Rendering with Shadow}

In this setting the camera and illumination are unchanged but the pose stems from a held-out set and hence varies significantly from the training set, leading to new shadow casts. As expected, existing methods (NPC, DANBO) overfit the training poses and cannot reproduce novel shadows. \figref{fig:qual_baselines} shows how for frames that had body parts casting shadows on other regions, our method produced more accurate shadows and albedo. 
\tabref{tab:baselines-novel-pose} quantifies the gains across novel poses and \tabref{tab:baselines-novel-pose-shadows} across the subset of the novel poses that has a shadow cast across the body. Improvements are consistent across all three metrics.

\subsection{Outdoor reconstruction}

Outdoor capture with strong directional light has not yet been attempted with neural body models.
In this challenging setting, all methods attain a lower quality since cameras are spaced further apart, and the 3D input pose used as input to the neural body models is less reliable, as estimated with off-the-shelf 2D pose detection and lifting methods. Nevertheless, Figure~\figref{fig:qual_baselines} shows how our model can accurately optimize the light direction and predict realistic shadows, including on the ground.
To map shadows to the ground, we estimate the ground plane from the reconstructed foot positions and cast the Gaussian shadow on it by modulating the static background with the ground shadow map.

\tabref{tab:baselines-novel-pose-shadows} shows that our method consistently improves the PSNR while the perceptual metrics SSIM \cite{wang2004image} improves only in one and the baselines perform better for LPIPS \cite{zhang2018perceptual}. This lower performance in perceptual metrics is expected because these metrics normalize for brightness and contrast differences, thereby lessening the importance of producing proper shading and shadowing. In addition, the texture and geometry detail of our method is slightly lower, which we attribute to the separation into shading and albedo imposing additional constrains, thereby leading to slightly less detailed reconstructions.

We also ran the official implementation of Relighting4D (R4D)~\cite{chen2022relighting} on this dataset, providing the same segmentation masks and SMPL body model as to our method (our method only uses the skeleton, not the surface). As the first stage of R4D is NeuralBody \cite{peng2021neural} that does not take shadowing into account, it produces dark floaters in the space to approximate the hard shadow, hindering their subsequent relighting module from estimating shadow and shading correctly. Figure~\figref{fig:qual_baselines} result showcases how important simultaneous optimization of shadow, shading, albedo, and geometry is in our method.

\subsection{Render Time Comparison}
\begin{table}[]
\centering
\resizebox{0.8\linewidth}{!}{
\begin{tabular}{|l|c|}
\hline
\multicolumn{1}{|c|}{\textbf{Method}}           & \textbf{render time {[}s{]}} \\ \hline
{DANBO + DS}               & { 17.13} \\ \hline
{DANBO + DS + GSC}         & { 17.47} \\ \hline
{DANBO + DS + NeRFSC}      &  21.4  \\ \hline \hline
{DANBO + DS + GSC-HDRi-8}  & { 20.70}  \\ \hline
{DANBO + DS + GSC-HDRi-16} & { 23.57} \\ \hline
{DANBO + DS + GSC-HDRi-32} & { 29.49} \\ \hline
{DANBO + DS + GSC-HDRi-64} & { 41.22} \\ \hline
\end{tabular}
}
\caption{\textbf{Render time.} The overhead of Gaussian Shadow Casting (GSC) is minimal on DANBO with diffuse shading (DANBO + DS) and enables casting many rays (64 for GSC-HDRi-64). By contrast, NeRF shadow casting (NeRFSC) doubles the runtime with every light source, making training prohibitively slow and HDRi relighting impractical.}
\label{tab:render time}
\end{table}

\tabref{tab:render time} lists the render time of our baseline compared to our full model. Casting shadows with GSC has minimal overhead (0.3s for one ray, only 2\% of the entire render time), enabling efficient training alongside NeRF optimization. Casting a shadow with the NeRF baseline requires processing twice the number of samples by the NeRF. The deferred shading model creates one occlusion ray and each of these secondary rays requires a similar number of samples as for the primary ray. Already with a single light source, this increases runtime by $25\%$, a ten-fold difference to GSC. 

\subsection{Relighting with Environment Maps}

The shadow computation not only benefits training time but also enables computing shadow maps for multiple light sources, including illumination by continuous environment maps. \figref{fig:teaser} shows relighting with two different HDRi maps (obtained from Poly Haven \cite{polyhaven}) by casting 64 secondary light rays towards the environment map for each pixel. One of those light rays is importance sampled, going towards the brightest region in the HDRi.
In both cases, the bright sun casts a strong shadow while the colored light from the environment leads to natural shading that matches the character with the environment. This enables placing the reconstructed characters into new environments and giving them a natural and consistent look with respect to the rest of the scene while still containing cast shadows.

\begin{table}[t]
\caption{\textbf{Ablation on Synthetic Sequence.} We evaluate on novel-poses on a training camera as well as novel-poses in a novel-view. 
}
\centering
\resizebox{\linewidth}{!}{
\setlength{\tabcolsep}{0pt}
\begin{tabular}{lcccccccccccccccccccccccc}
\toprule
& \multicolumn{3}{c}{Novel Pose} & \multicolumn{3}{c}{Novel View}\\
\cmidrule(lr){2-4}\cmidrule(lr){5-7}%
  & PSNR$\uparrow$  & SSIM$\uparrow$  & LPIPS$\downarrow$
   & ~PSNR\phantom{$\uparrow$}& SSIM\phantom{$\uparrow$}  & LPIPS\phantom{$\uparrow$}
\\
\rowcolor{Gray}
GT Light
& 21.23& \textbf{0.830}& 0.184
& 26.68& \textbf{0.895}& 0.155\\
Detached Normals
& 20.43& 0.814& 0.192
& 26.18& \underline{0.893}& 0.158\\
\rowcolor{Gray}
No Diffuse
& 21.22& 0.802& 0.178
& 25.08& 0.867& 0.163\\
Ours
& \textbf{22.30}& \underline{0.827}& \textbf{0.176}
& \textbf{27.32}& 0.882& \textbf{0.154}\\
\end{tabular}
\label{tab:ablations}
}
\end{table}
\begin{figure*}[t!]
\centering
\includegraphics[width=0.8\linewidth]{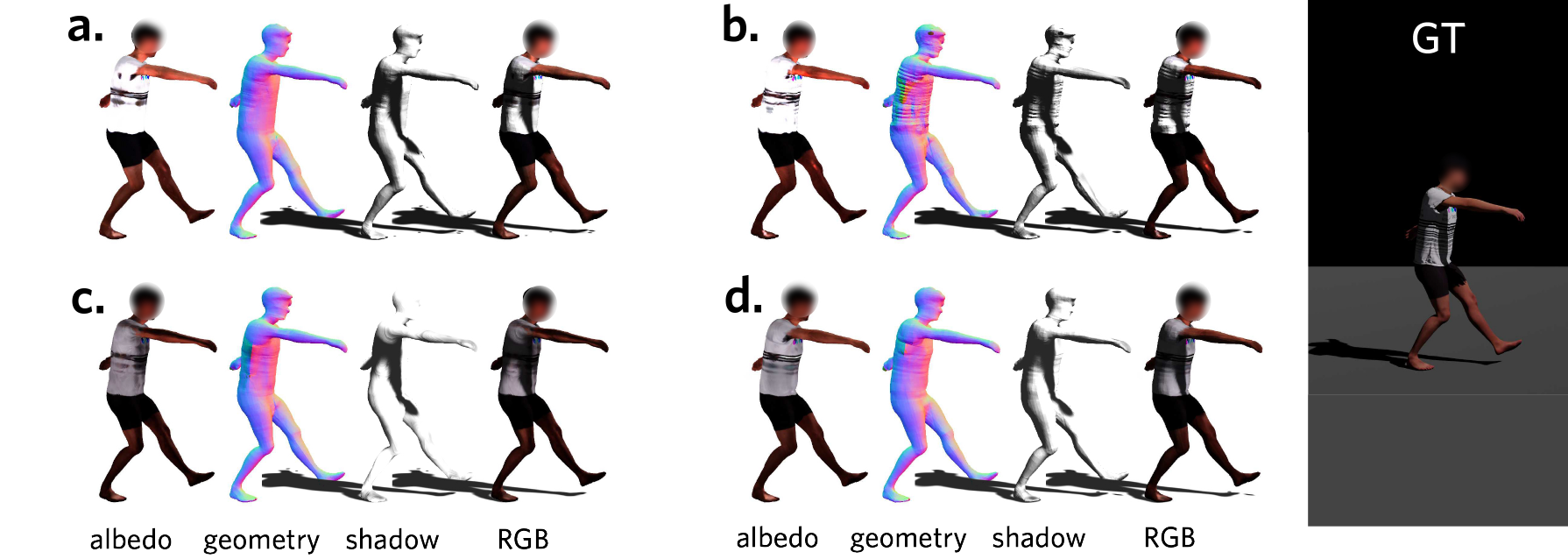}
\caption{\textbf{Ablation Comparisons (novel-pose).} a) The model trained with the groundtruth light direction. b) The model trained while detaching the gradients from the surface normals during diffuse shading. c) The model trained without diffuse shading. d) Our full model.}
\label{fig:ablations}
\end{figure*}

\subsection{Ablation Study}
We test a variety of implementation details in our model, including using just the Gaussian density model to cast shadows without a diffuse shading component, not optimizing the light source and instead using ground truth lighting, and finally detaching the normals during the diffuse shading computation. The results of which can be seen in ~\figref{fig:ablations}. \tabref{tab:ablations} shows that each of our contributions improves reconstruction quality.

\parag{Diffuse Shading.}
Shadowing alone does not account for accurate shading based on how 
incident the light hits the surface. Moreover, 
the Gaussians 
cast long-range shadows,
but their smooth and low-resolution approximation to the NeRF's density prevents them from representing finer details such as small extremities (nose, fingers). Backwards-facing areas are however handled by the diffuse model (negative light-normal angle that is clamped to zero and thereby also scales the light contribution to 0). As a result, 
~\figref{fig:ablations}c shows how disabling the shading misses these details and bakes some shadows into the generated albedo maps, which does not happen with our full model.

\parag{Light Optimization.}
Our method is able to fit the direction of the light source and the ambient intensity. %
We observe accurate light recovery when the light is initialized randomly, e.g. when coming from the back the angle error is only 1.36 degrees. To showcase the robustness and generality, the real sequences start with the light coming straight down and the synthetic one even set opposite to the true direction. We found providing the ground truth light direction did not improve results (comparing \figref{fig:ablations}a and d).

\parag{Detached Normals.}
Finally, we compare results between a model where network gradients could backpropagate through the surface normals used in the diffuse shading, ~\figref{fig:ablations}b, to see whether or not artifacts in the shading could smoothen out the geometry. We find that the surface is indeed affected by the gradients backpropagating through the diffuse computation and observe more faithful geometry reconstruction in~\figref{fig:ablations}d.

\subsection{Limitations}

The Gaussian cast shadows model long-range effects, such as the arm casting a shadow on the leg but the smooth Gaussians lack high frequency details. This is a minor drawback since the diffuse shading already faithfully reproduces the light intensity fall-off as the light direction becomes more incident with the surface and therefore shades the back side of small extremities (i.e. nose and fingers) well. A future extension could be to integrate mid-scale effects with screen-space ambient occlusion and shading.

Moreover, we noticed that disentangling color into shading and albedo, compared to the original DANBO backbone, leads to slightly lower image reconstruction metrics when shading effects are minimal. We attribute this to the additional constraints that are imposed on the model. However, the overall performance in novel light conditions is still improved significantly by our model.

\section{Conclusion}
We enabled the 3D reconstruction of human motions in uncontrolled environments
by a Gaussian shadow model that applies to dynamic scenes and is differentiable for iterative refinement. 
The reconstructed characters support reposing and relighting in novel environments. They are equipped with global shadow computation, diffuse shading, geometric reconstruction, and a consistent albedo, much like hand-crafted computer graphics models would provide.

{\small
\bibliographystyle{ieee_fullname}
\bibliography{egbib}
}

\appendix
This supplemental document supplies additional details on the mathematical derivation and training details to aid future work and possible extensions. It also contains extra figures showcasing more results for the experiments described in the main document, as well as demonstrating our relighting methods on other datasets. 

\section{HDRi Relighting}
\label{sec:relighting}
We are able to relight, not only by changing a single, primary light source direction, but by using a high dynamic range image (HDRi) that defines the environment illumination. For HDRi relighting, we rely on the Gaussian density model to query visibility and pair it with a diffuse reflectance model. We cast multiple secondary rays for each pixel from the surface of the model towards to the environment map, typically $64$ secondary rays per pixel. The first of which is important sampled towards the brightest region in the HDRi, i.e. the sun. 
The rest are sampled according to diffuse reflection, with probability proportional to the cosine angle between normal and sample direction. This distribution can be attained by setting the ray direction as the unit surface normal and adding a random point on the unit sphere.
This is a simple yet effective model that serves our purpose by sampling rays that contribute most with high likelihood, considering both the brightest region in the environment map and those that contribute most to diffuse reflection. In the same manner, the Gaussian shadow casting could also be integrated into a physically accurate illumination model by including a full BRDF function and an unbiased importance sampling method for reducing variance.

Relighting using HDRis can be seen in the Supplemental Video and the teaser in the main paper.

\begin{figure}[t]
\centering
\includegraphics[width=1.0\linewidth]{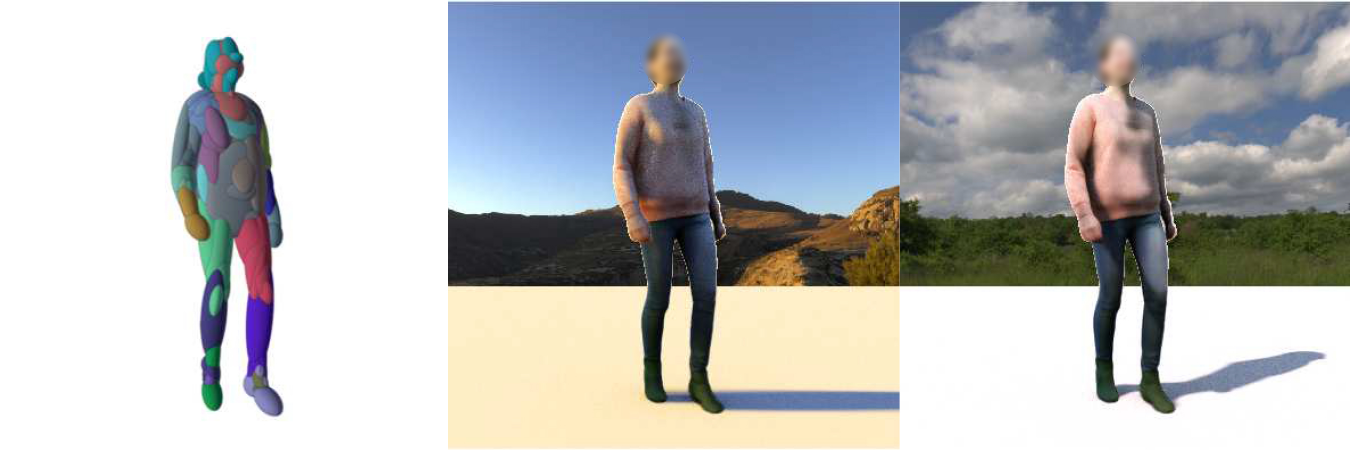}
\caption{\textbf{HDRi Relighting on MonoPerfCap.} 
Our Gaussian relighting method can work on recordings done under uniform illumination, even monocular datasets.
}
\label{fig:nadia}
\end{figure}

\paragraph{Relighting on MonoPerfCap and Animal Datasets.}
Our Gaussian method can also be used as a standalone relighting tool for datasets captured under uniform illuminations. These datasets have very few lighting and shading effects which allows for the direct interpretation of the neural color field as the albedo. We can simply fit the Gaussian density model and relight the learned avatars with HDRis. We test this paradigm on a monocular sequence from the MonoPerfCap dataset \cite{xu2018monoperfcap}. See \figref{fig:nadia} for the results.

\begin{figure}[t]
\centering
\includegraphics[width=1.0\linewidth]{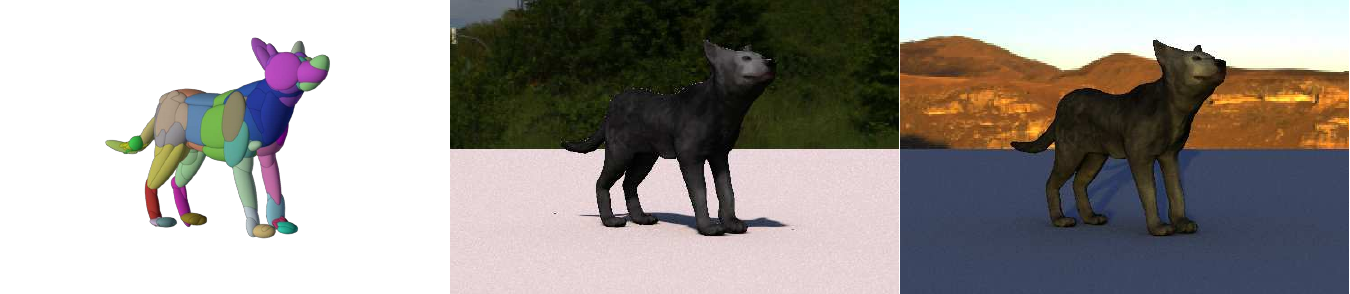}
\caption{\textbf{HDRi Relighting on Animal Dataset.} 
Our method can extend beyond human avatars as we do not necessitate any templates.
}
\label{fig:wolf}
\end{figure}

Likewise, due to the template-less nature of our implementation, we are able to learn bodies with Gaussian density models for non-human characters. We test this using the Animal dataset \cite{li2022tava} as seen in \figref{fig:wolf}.

\section{Novel Poses on Outdoor Sequence}
\begin{figure*}[t!]
\centering
\includegraphics[width=1.0\linewidth]{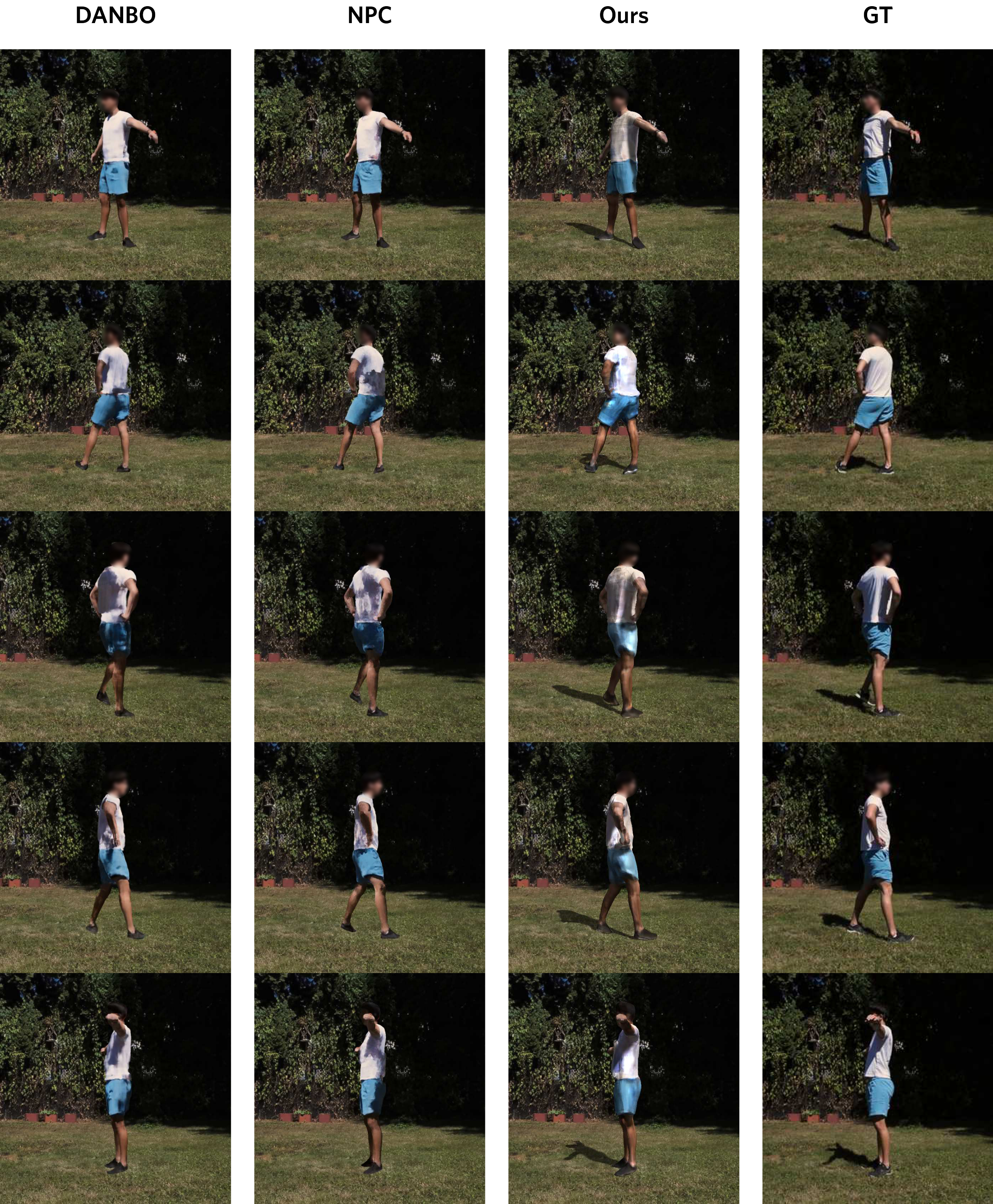}
\caption{\textbf{Outdoors in sunlight (Real Sequence 1).} 
Novel poses rendered using DANBO \cite{su2022danbo}, NPC \cite{su2023npc} and our method. Our method has more consistent lighting and shadows whereas the baselines suffer from large shadow changes from small pose variations.
}
\label{fig:supp_qual}
\end{figure*}

We showcase more novel-pose results on a real sequence captured outdoors in bright daylight \figref{fig:supp_qual}. Our explicit lighting module results in more accurate shadows compared to the baselines. The baselines, which overfit to the training set, are highly inconsistent with small perturbations in pose leading to large changes in the shadow.

\begin{figure}[t]
\centering
\includegraphics[width=1.0\linewidth]{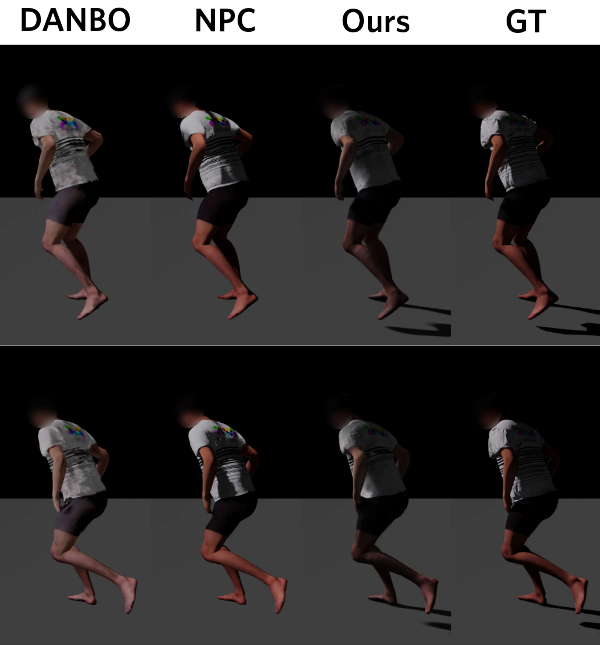}
\caption{\textbf{Novel-view Synthesis (Synthetic Sequence).} Due to the explicit nature of our lighting module, our models are robust to changes in view. The baselines are highly pose-dependant with small changes in pose affecting the image drastically.
}
\label{fig:novelview}
\end{figure}

\section{Novel-View Rendering Results}
We also test the ability to render novel views using our method. We find that due to the explicit nature of our shading computations, the lighting and shadows are still accurate. The baselines fail to produce accurate illuminations and are much more susceptible to small pose variations leading to large changes, as seen in ~\figref{fig:novelview}. In contrast to the training cameras, the novel view is backlit with large parts of the trunk in shadow. Consequently, the quantitative analysis in \tabref{tab:novelview} demonstrates even larger performance gains than the novel pose evaluation in the main document. The result further highlights the importance of our explicit shadow-casting method.
\begin{table}[b]
\caption{\textbf{Novel-pose in novel-view synthesis (all test frames).} As our method is explicit, large changes is view direction still result in accurate shadowing unlike the baselines.
}
\centering
\resizebox{1\linewidth}{!}{
\setlength{\tabcolsep}{10pt}
\begin{tabular}{lcccccccccccccccccccccccc}
\toprule
& \multicolumn{3}{c}{Novel View}\\
\cmidrule(lr){2-4}%
  & PSNR$\uparrow$  & SSIM$\uparrow$  & LPIPS$\downarrow$
\\
\rowcolor{Gray}
DANBO
& 18.85& 0.773& 0.179\\
NPC
& \underline{20.64}& \underline{0.823}& \underline{0.157}\\
\rowcolor{Gray}
Ours
& \textbf{27.19}& \textbf{0.873}& \textbf{0.153}\\
\end{tabular}
\label{tab:novelview}
}
\end{table}

\section{Training Details}
\subsection{Loss Functions \& Regularizations}
\label{sec:losses}
Our primary objective is the accurate reconstruction of the neural character renders $\hat{\mI}$ and the training images $\mI$. We use a standard photometric reconstruction loss between the pixel color values in the training image, $\pixelcoltrain$, and reconstruction, $\pixelcol$.

\begin{equation}
\label{eq:rgbloss}
\RGBloss = |\pixelcol - \pixelcoltrain|
\end{equation}

We augment this by also using a mask loss that operates between the integrated accumulation, $\accumulation$, and the foreground mask, $\mask$.

\begin{equation}
\label{eq:maskloss}
\Maskloss = |\accumulation - \mask|
\end{equation}

Our final photometric loss is one we dub the grey loss. Its objective is to initialize the RGB head of the NeRF to output a light grey value such that when the standard RGB loss starts to have an influence on the training, it is not prone to getting stuck in a local minimum with shadows already learnt caused by an initialization resulting in darker color values. It also provides enough time for the light direction to optimize and fit before the NeRF overfits to the shadows as we interpolate between the grey loss and the RGB loss.

\begin{equation}
\label{eq:rgbloss}
\Greyloss = |\pixelcol - 0.75|
\end{equation}

To fit the Gaussians, we introduce a Gaussian Density loss which minimizes the squared distance between the Gaussian density function at a given query location, $\mG(\vx)$, and the density head of the NeRF at the same query location, $\mD(\vx)$.

\begin{equation}
\label{eq:gaussiandensityloss}
\GaussianDloss = ||\mG(\vx) - \mD(\vx)||^2
\end{equation}

\begin{figure}[h]
\centering
\includegraphics[width=1.0\linewidth]{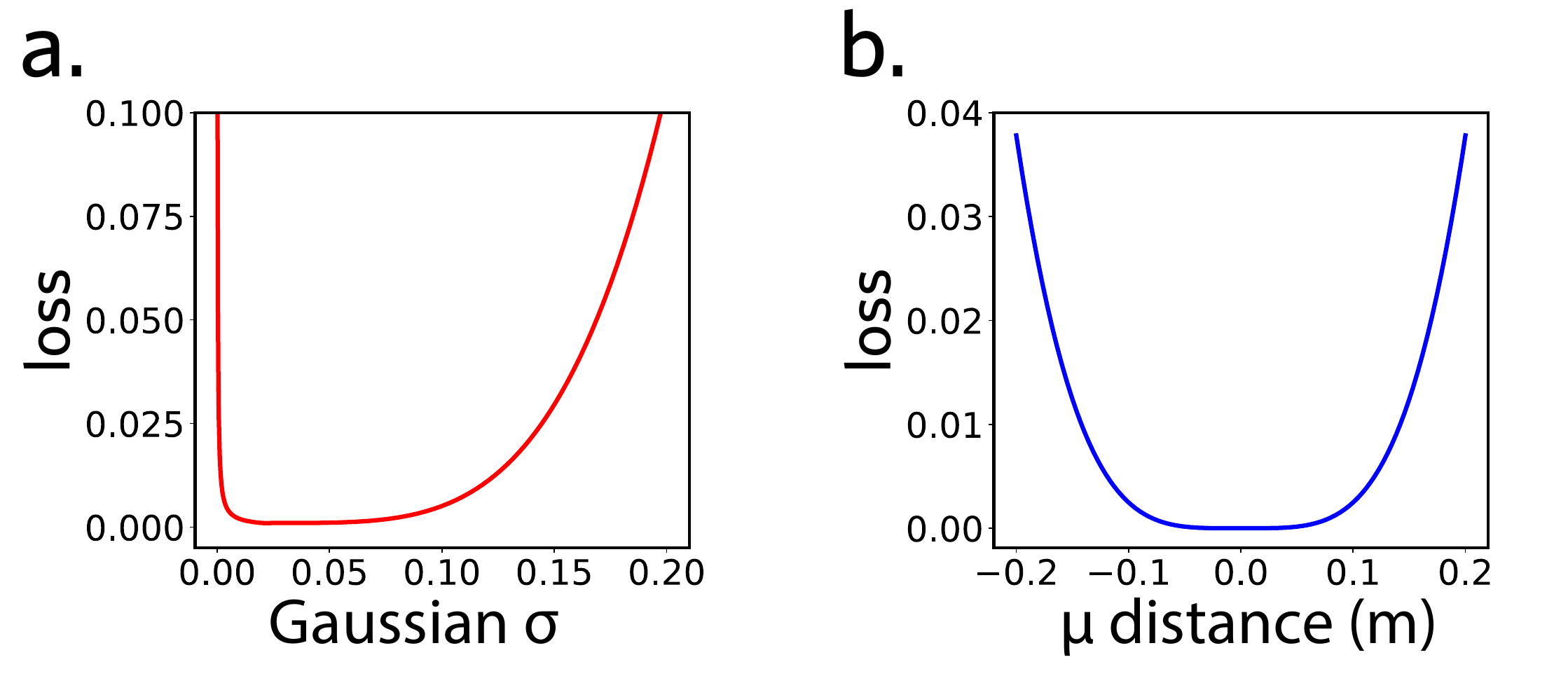}
\caption{\textbf{Regularization on Gaussian Density Model.} 
a) The regularization function for the standard deviations of the Gaussians, constraining the size. b) The regularization function for the means of the Gaussians, keeping the Gaussians close to the center of the bones. 
}
\label{fig:gaussianregularization}
\end{figure}

We regularize our Gaussians by supervising their mean and standard deviations. The standard deviation regularization limits the size of the Gaussians to approximately be within $2.5$ and $50$ centimeters, while the mean regularization prevents Gaussians from drifting too far from the bone centers, $\vb$. These loss functions are visualized in Figure \ref{fig:gaussianregularization}.

\begin{equation}
\label{eq:gaussiansigmaloss}
\GaussianSloss = \begin{cases} 
      \frac{2e-5}{\sigma} & \sigma \leq 0.02 \\
      100(\sigma-0.02)^4 + 0.001 & \sigma > 0.02 
   \end{cases}
\end{equation}

\begin{equation}
\label{eq:gaussianmeanloss}
\GaussianUloss = (100(\mu - \vb)^4 + 1)^\frac{1}{4} - 1
\end{equation}

We also regularize the ambient intensity, $\ambient$, to be somewhat dark values to prevent the model from setting a bright ambient value and learning all of the shadows as part of the NeRF color.

\begin{equation}
\label{eq:ambienloss}
\Ambientloss = ||\ambient - 0.1||^2
\end{equation}

We adopt the SDF-based density field from VolSDF~\cite{yariv2021volume} for improved normals, and regularize the SDF network using an Eikonal loss~\cite{gropp2020igr}
$\Eikonalloss$ to predict proper level sets,
\begin{equation}
\Eikonalloss = ||\hat{\normal}(\vx) - 1||^2_2
\end{equation}
and use a curvature loss, $\Curvatureloss$, to smoothen out the geometry by minimizing the difference between neighbouring normals,

\begin{equation}
\label{eq:curvatureloss}
\Curvatureloss = ||\hat{\normal}(\vx) - \hat{\normal}(\vx + \epsilon)||^2,
\end{equation}
with $\epsilon$ a small random perturbation.

Our total loss is the sum of all of these losses and regularization terms, each with a weighing function, $\alpha_i(t)$, for loss term $i$ and training iteration $t$.

\subsection{Scheduled Learning}
\label{sec:learning}
Our scheduled learning can be split into $3$ segments. Denoting a change in the weights for each of the loss terms throughout the training using linear interpolation.

\paragraph{Segment 1: Density Fitting $\sim{1k}$ Iterations.} In this step, the main goal for the model is to train the neural field's density to fit the silhouette of the character. It begins with high weights for only $\Maskloss$ and $\Greyloss$ alongside the regularizers for curvature, $\Curvatureloss$, and Eikonal constraints $\Eikonalloss$.  We need this first stage as our shading computations rely on accurate depth maps, normal maps and accurate Gaussian fits, which the latter requires an accurate density field to fit to. Training the RGB head directly from the start results in many artifacts that the network cannot recover from due to the deferred nature of our shading computations.

\paragraph{Segment 2: Gaussian Density Model Fitting $\sim{4k}$ Iter.}
This segment marks the addition of the Gaussian density loss, $\GaussianDloss$, and its regularizers, $\GaussianUloss$ and $\GaussianSloss$. At which point the parameters of the Gaussians, $\mG$, are optimized to fit to the pretrained NeRF's density.

\paragraph{Segment 3: Light Fitting \& RGB Fitting}
This step switches from using the grey loss to the RGB loss. Iterpolation between the two ensures a smooth transition between purely optimizing for the silhouette and our target color reconstruction. In our experiments, $1k$ iterations were sufficient to fully optimize the light direction, at which point the diffuse and Gaussian shadow computation is fairly accurate, allowing the neural color field to learn color without shadows, more closely resembling the albedo, see the Supplemental Video.

\begin{figure}[h]
\centering
\includegraphics[width=1.0\linewidth]{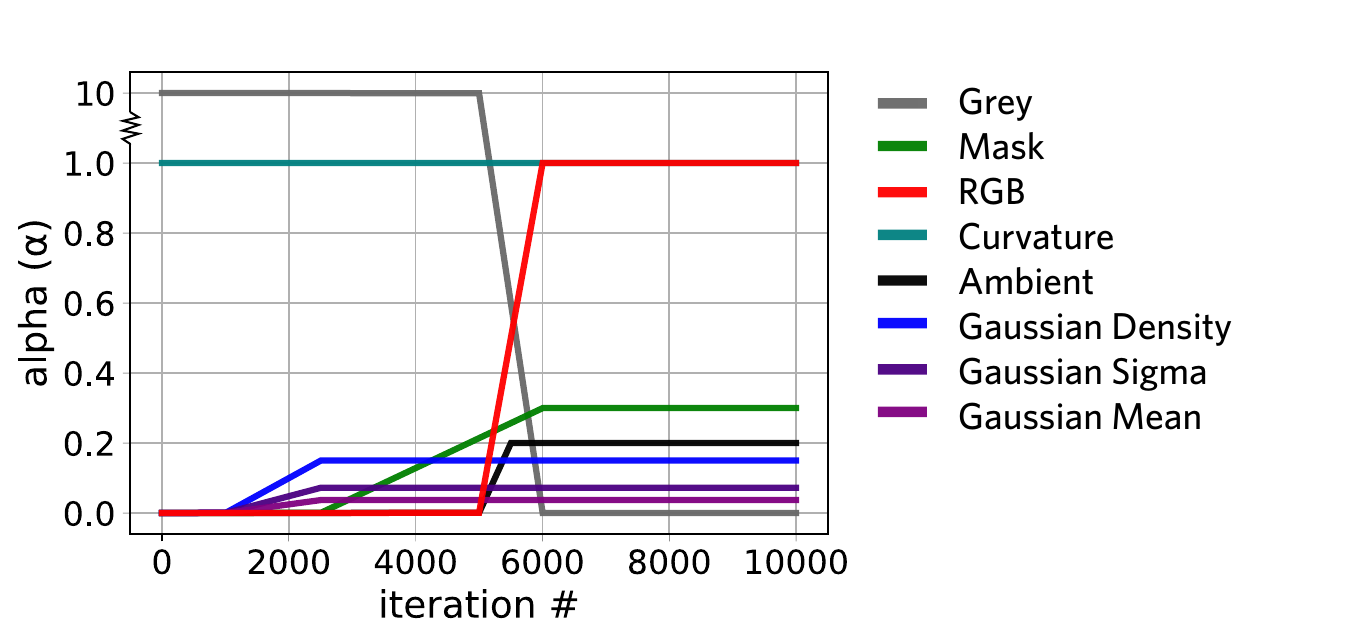}
\caption{\textbf{Weight Modulation.} 
How the weights $\alpha_i(t)$ change throughout the training. First focusing on fitting a grey silhouette and the Gaussian model, later transitioning to fir the RGB alongside ambient regularization.
}
\label{fig:weights}
\end{figure}

\paragraph{Weight Modulation:}
As previously mentioned, our total loss is the sum of all of our loss terms each with a weighing function $\alpha_i(t)$ which modulates the weight during training to allow the previously mentioned stages to train properly. We plot the value of each of the weighing functions over the training iterations in \figref{fig:weights}. They are linearly interpolating between two values over a number of iterations with a hold-off period.

\section{Analytical Gaussian Integral}
\label{sec:gaussian-analytic}
Our goal is to derive a 1D function $G^r(t)$ that represents the density along a ray $\vr$ that we can integrate to acquire the cast shadow. We start our derivation from the 3D anisotropic Gaussian $\mG(\vx)$ that we use to approximate the body density field,
\begin{align}
    \mG(\vx)&=\cC\exp\left[{-0.5(\vmu-\vx)^T\Sigma^{-1}(\vmu-\vx)}\right] \nonumber\\
    &=\cC\exp\left[-0.5(\vmu^T\Sigma^{-1}\vmu-2\vmu^T\Sigma^{-1}\vx + \vx^T\Sigma^{-1}\vx)\right]. 
    \label{eq:gaussian-1d-rewrite-0}
\end{align}
We can infer the density along the ray, $G^r(t)$, by parameterizing the 3D positions along the ray by the distance $t$ from the origin $\vr_o$, with $\vx=\vr_o+t\cdot\vr_d$, and substituting it into \feqref{eq:gaussian-1d-rewrite-0},
\begin{align}
    &G^r(t)\nonumber\\
    &=\hat{\cC}\exp\left[-0.5(t^2\vr_d^T\Sigma^{-1}\vr_d - 2t\vr_d^{T}\Sigma^{-1}(\vr_o-\vmu))\right] \nonumber\\
    &= \hat{\cC}\exp\big[-0.5\vr_d^T\Sigma^{-1}\vr_d\left(t^2 - 2t\frac{\vr_d^{T}\Sigma^{-1}(\vr_o-\vmu)}{\vr_d^T\Sigma^{-1}\vr_d}\right)\big],
\label{eq:gaussian-1d-rewrite-1} 
\end{align}
where %
$\vr_d$ is the ray direction, and 
$\hat{\cC}$ consists of the terms that are constants with respect to the ray distance $t$, separated by the equality $\exp(a+b) = \exp(a) \exp(b)$,
\begin{align}
\hat{\cC} = \cC\exp\left[-0.5(\vmu-\vr_o)^T\Sigma^{-1}(\vmu-\vr_o)\right].    
\end{align}
We then reorganize \feqref{eq:gaussian-1d-rewrite-1} by substituting $\hat{\sigma}=\frac{1}{\sqrt{\vr_d^T\Sigma^{-1}\vr}}$ and $\bar{\vmu}=\frac{\vr_d^{T}\Sigma^{-1}(\vr_o-\vmu)}{\vr_d^T\Sigma^{-1}\vr_d}$, 
\begin{align}
        G^r(t)&=\hat{\cC}\exp\big[-0.5\frac{(t^2 - 2t\bar{\vmu})}{\bar{\sigma}^2}\big]\nonumber\\
        &=\hat{\cC}\exp\big[-0.5\frac{(t^2 - 2t\bar{\vmu} + \bar{\vmu}^2) - \bar{\vmu}^2}{\bar{\sigma}^2}\big]\nonumber\\
        &=\hat{\cC}\exp\big[-0.5\frac{(t - \bar{\mu})^2  - \bar{\mu}^2}{\bar{\sigma}^2}\big]\nonumber\\
        &=\bar{\cC}\exp\left[-\frac{(\bar{\mu} - t)^2}{2\bar{\sigma}^2}\right],\label{eq:gaussian-1d-final}%
\end{align}
where, again, $\bar{\cC}$ absorbs the terms that are constant to $t$,
\begin{equation}
     \bar{\cC}= \cC\exp\left[-0.5\left((\vmu-\vr_o)^T\Sigma^{-1}(\vmu-\vr_o) - \frac{\bar{\vmu}^2}{\bar{\sigma}^2}\right)\right],
\end{equation}
and we arrive at $G^r(t)$ that assumes the form of a 1D Gaussian density function with mean $\bar{\mu}$, std $\bar{\sigma}$, and scaling factor $\bar{\cC}$. As the integral through a Gaussian can be computed in closed form through the error function, this enables analytical integration of the density along the ray, which in turn enables our method to cast shadows efficiently.

\end{document}